\pdfoutput=1

\documentclass[11pt]{article}
\usepackage[table,xcdraw]{xcolor}
\usepackage{acl}
\usepackage{caption}
\usepackage{subcaption}

\usepackage{xspace}
\usepackage{booktabs}
\usepackage{graphicx}
\usepackage{multirow}
\usepackage{tikz}
\usetikzlibrary{plotmarks}
\usepackage{fontawesome5}
\usepackage{tabularx}
\usepackage{bbm}
\usepackage{amsfonts}
\usepackage{amsmath}
\usepackage{amssymb}
\usepackage{algorithm}
\usepackage{algpseudocode}
\usepackage[normalem]{ulem}
\usepackage{times}
\usepackage{latexsym}
\usepackage{pifont}
\newcommand{\cmark}{\ding{51}}
\newcommand{\xmark}{\ding{55}}
\usepackage{mathtools} 
\usepackage{extarrows}
\usepackage{tabularx}
\usepackage{cleveref}
\crefformat{section}{\S#2#1#3} 
\crefformat{subsection}{\S#2#1#3}
\crefformat{subsubsection}{\S#2#1#3}

\usepackage[T1]{fontenc}

\usepackage[utf8]{inputenc}

\usepackage{microtype}
\usepackage{amsmath}
\usepackage{inconsolata}

\usepackage{soul}
\usepackage{enumitem}
\usepackage{todonotes}
\usepackage{verbatim}
\usepackage{relsize}
\usepackage{array}
\newcolumntype{?}{!{\vrule width 1pt}}
\definecolor{darkgreen}{rgb}{0.0, 0.5, 0.0}
\definecolor{carnelian}{rgb}{0.7, 0.11, 0.11}
\newcommand{\halfcheckmark}{\cmark\makebox[-1pt][r]{\raisebox{2pt}{\scalebox{0.5}[1]{\normalfont\symbol{'26}}}}}

\usepackage{pict2e}
\usepackage{amssymb}

\usepackage{listings}
\newcommand{\ComplexChain}{Complex Chain}
\newcommand{\SimpleChain}{Simple Chain}
\newcommand{\NoChain}{No Chain}
\newcommand{\PA}{Paraphrase}
\newcommand{\EDIT}{Edit}

\title{Detecting Machine-Generated Long-Form Content with Latent-Space Variables}

 \author{ {Yufei Tian}\quad  \textbf{Zeyu Pan} 
\quad \textbf{Nanyun Peng} \\[7pt]
         University of California, Los Angeles\\[3pt]
         {
         \texttt{yufeit@g.ucla.edu} \quad 
         }
         }

\begin{document}
\maketitle
\begin{abstract}
The increasing capability of large language models (LLMs) to generate fluent long-form texts is presenting new challenges in distinguishing machine-generated outputs from human-written ones, which is crucial for ensuring authenticity and trustworthiness of expressions. Existing zero-shot detectors primarily focused on token-level distributions, which are vulnerable to real-world domain shifts including different prompting and decoding strategies, and adversarial attacks. We propose a more robust method that incorporates abstract elements---such as \textit{event transitions}---as key deciding factors to detect machine vs. human texts, 
by training a latent-space model on sequences of events or topics derived from human-written texts.
On three different domains, 
machine generations which are originally inseparable from humans' on the token level can be better distinguished with our latent-space model, leading to a 31\% improvement over strong baselines such as DetectGPT \cite{mitchell2023detectgpt,bao2023fastdetectgpt}. Our analysis further  reveals that, unlike humans, modern LLMs like GPT-4 generate event triggers and their transitions differently, an inherent disparity that help our method to robustly detect machine-generated texts. 







\end{abstract}

\section{Introduction}\label{sec:intro}

In today's digital world, large language models (LLMs) such as GPT-4 have transformed various daily tasks with their human-like text generation capability, such as drafting emails and essays. However, their potential misuse poses substantial risks including impersonation, misinformation, and academic dishonesty \cite{detect_survey1}. This highlights the need for effective detection mechanisms. Existing AI content detectors can be categorized into 1) a priori methods such as watermarking \cite{watermark}, 2)  parameterized methods such as fine-tuned classifiers \cite{hu2023radar}, and 3) zero-shot methods that rely on certain statistical differences \cite{vasilatos2023howkgpt,mitchell2023detectgpt,bao2023fastdetectgpt}. This paper focuses on the last due to its general pertinence in practice: end users may still choose non-watermarked LLMs outside the distribution of the fine-tuned classifiers \cite{detect_survey2,ghosal2023survey}.

\begin{figure}[!t]
    \centering
    \includegraphics[width=1\linewidth]{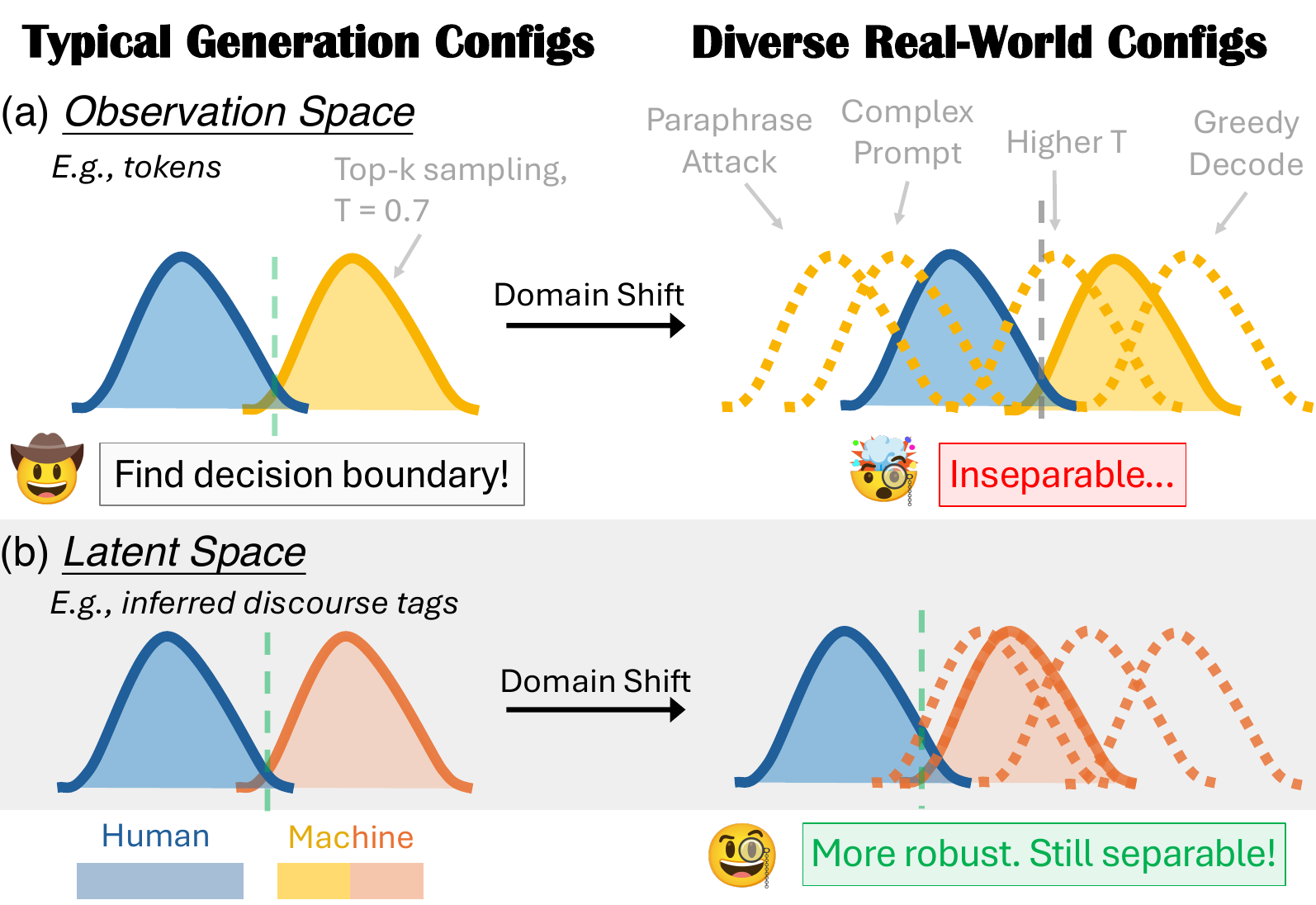}
    \caption{(a) Existing zero-shot detectors that rely on the token distributions (observation space statistics) are not robust to various real-world scenarios such as high decoding temperature, complex prompts, and adversarial attacks. (b) Our detector with latent features (\textit{e.g.,} discourse tags) are more robust to these changes. 
    } \vspace{-3mm}
    \label{fig:teaser}
\end{figure}

Existing zero-shot methods to distinguish machine-generated texts (\textbf{MGTs}) from human-written texts (\textbf{HWTs}) typically assume an \textit{unchanging} relationship between machine and human outputs, ignoring potential \textit{distribution shifts} resulted from changes in generation setups \cite{zellers2019defending_fake_news} or adversarial attacks \cite{shi2024redteam,wang2024stumbling}. 
As is shown in Figure \ref{fig:teaser}(a), prior zero-shot detection methods usually assume that MGTs consistently exhibit higher log-likelihood (or conditional curvature) than HWTs. 
However, 
certain changes---including increased decoding temperature, paraphrasing, and word substitution---can alter the distribution of MGTs \cite{binoculars}, making MGTs and HWTs \textit{\textbf{inseparable}} by log-likelihood after domain shift. In this work, we further identify that complex prompts (\textit{e.g.,} ask to prepare an outline before generation) 
can alter the typical relationship between MGTs and HWTs. Consequently, when considering MGTs from diverse sources in real world, existing detectors become less effective.

Towards more robust detection, we aim to answer the research question: 
\textit{Is there a shared feature among all MGTs, regardless of the generation configurations and adversarial attacks?} \citet{wang2022language_stochastic,deng2022model_criticism,tian2024large} argue that machine-generated long texts are fluent at word level, but lack discourse coherence 
in terms of \textit{topic} or \textit{event transitions}.
We hence hypothesize that MGTs and HWTs are more separable in such \textbf{\textit{latent space}} illustrated in Figure \ref{fig:teaser}(b), because those underlying features can not be easily captured by next token probabilities (\textit{i.e.,} optimization in observation\footnote{We use \textit{observation}, \textit{sample}, and \textit{token} interchangeably throughout this paper.} space). In addition, paraphrase or edit attacks change original texts with semantically similar yet lexically different alternatives. While they effectively alter token-level distributions, the high-level, hidden representations remain similar.  

We test our hypothesis on three writing tasks: creative scripts, news, and academic essays. 
We explored latent variables 
like part-of-speech tags, topics, verbs, and event sequences.  Specifically, we first train a lightweight model (62M transformer, half the size of gpt-2-small) on the latent variables inferred from HWTs, and then compare the latent distributions between HWTs and MGTs 
at test time. We find detectors in observation and latent space exhibit complementary strengths ($\S$ \ref{subsec:combining}) in differentiating MGTs from various configurations. Integrating the criteria from both types yields the best performance.

We explore five features to represent the underlying structure, and find that \textbf{\textit{event trigger}} derived from information extraction models \cite{omnievent} is the most effective in separating HWTs from MGTs, outperforming strong token-space detector \cite{bao2023fastdetectgpt} by 31\% in AUROC. 
Our analysis in $\S$ \ref{sec:analysis} further reveals that LLMs such as GPT-4 exhibit a different preference from human in choosing event triggers (for creative writing) and event transitions (for news and science), 
and such a disparity cannot be bridged through explicit planning of these latent structures.

To sum up, we identify key factors that deceive existing detectors in real-world scenarios. We then demonstrate a significant discrepancy in hidden structures between current LLMs and humans, especially the selection and transitions of event triggers. Building on these insights, we propose a novel detection framework that employs latent variables to robustly differentiate between human- and machine- generated texts.

\section{Preliminary: Fragility of Existing Zero-Shot Detectors}\label{sec:fragility}

In this section, we introduce two popular lines of zero-shot detection methods (logit-based and perturbation-based), and then illustrate how they are fragile to manipulations in decoding, variations in prompting style, and adversarial attacks.

\subsection{Existing Detectors}

\paragraph{Logit-Based}
Logit-based methods commonly employ probability metrics of tokens. \citet{ppl_first} established a strong baseline for detecting machine-generated text through the average log probability under the generative model. The intuition behind is that language model text generation is auto-regressive; the model selects tokens based on relatively higher probability at each decision point, resulting in MGT exhibiting a markedly higher average log probability compared to HWT, which becomes the foundational assumption of perplexity-based detectors \cite{vasilatos2023howkgpt, xu2024detecting_ppl} and rank-based detectors \cite{su2023detectllm}, etc.

\paragraph{Perturbation-Based} Another notable hypothesis introduced by \citet{mitchell2023detectgpt} posits and verifies that MGT tends to occur in regions of negative curvature within the language model’s log probability function. Specifically, minor edits to MGT—referred to as perturbations—typically lead to a lower log probability under the model than the original text, whereas such rewrites of HWT may result in either higher or lower log probabilities.
\citet{bao2023fastdetectgpt} then increases its efficiency and efficacy by utilizing dual models that share the same tokenizer to expedite the perturbation process. Given text sample $x$ and scoring model $p_{\theta}$, conditional probability \textbf{\textit{curvature}} is defined as:
\begin{equation}
\vspace{-2mm}
\begin{aligned}
& \mathbf{d}\left(x, p_\theta, q_{\varphi}\right)=\frac{\log p_\theta(x \mid x)-\tilde{\mu}}{\tilde{\sigma}} 
\text{ where, }
\end{aligned}
\end{equation}
\begin{equation}
\begin{aligned}
& \tilde{\mu}=\mathbb{E}_{\tilde{x} \sim q_{\varphi}(\tilde{x} \mid x)}\left[\log p_\theta(\tilde{x} \mid x)\right] \\ 
& \tilde{\sigma}^2=\mathbb{E}_{\tilde{x} \sim q_{\varphi}(\tilde{x} \mid x)}\left[\left(\log p_\theta(\tilde{x} \mid x)-\tilde{\mu}\right)^2\right]\text{.}
\end{aligned}
\end{equation}
$\tilde{\mu}$ denotes the expected score of samples $\tilde{x}$ generated by the sampling model $q_{\varphi}$, and $\tilde{\sigma}^2$ is the expected variance of the scores.

\subsection{Influential factors for detection}

\begin{table}[]
\centering
\small
\setlength{\tabcolsep}{3pt}
\begin{tabular}{@{}llcc@{}}
\toprule
\multicolumn{2}{c}{\textbf{Method for Generation}} & \textbf{Logit Based} & \textbf{Pert. Based} \\ \midrule
Default & \begin{tabular}[c]{@{}l@{}}T=0.7\\ Direct Generation\end{tabular} & \color{darkgreen}{\cmark}                          & \color{darkgreen}{\cmark}                        \\
\rowcolor[HTML]{EFEFEF} Decoding                   & \begin{tabular}[c]{@{}l@{}}T=1.0\\ Direct Generation\end{tabular}        & \halfcheckmark  & \color{darkgreen}{\cmark}  \\
\multirow{2}{*}{Prompting} & Simple             & \halfcheckmark  & \halfcheckmark  \\
                           & Complex      & \color{red}{\xmark} & \color{red}{\xmark} \\
\rowcolor[HTML]{EFEFEF} 
\cellcolor[HTML]{EFEFEF}    & Paraphrase        & \color{red}{\xmark} & \color{red}{\xmark} \\
\rowcolor[HTML]{EFEFEF} 
\multirow{-2}{*}{\cellcolor[HTML]{EFEFEF}Attack} & Edit                                                        & \color{red}{\xmark}                      & \color{red}{\xmark}                    \\ \bottomrule
\end{tabular}
\caption{Different methods for generation 
and whether existing zero-shot detectors are robust to them.}
\label{tab:attack_table}
\end{table}

\begin{figure}[!t]
    \centering
    \includegraphics[width=1\linewidth]{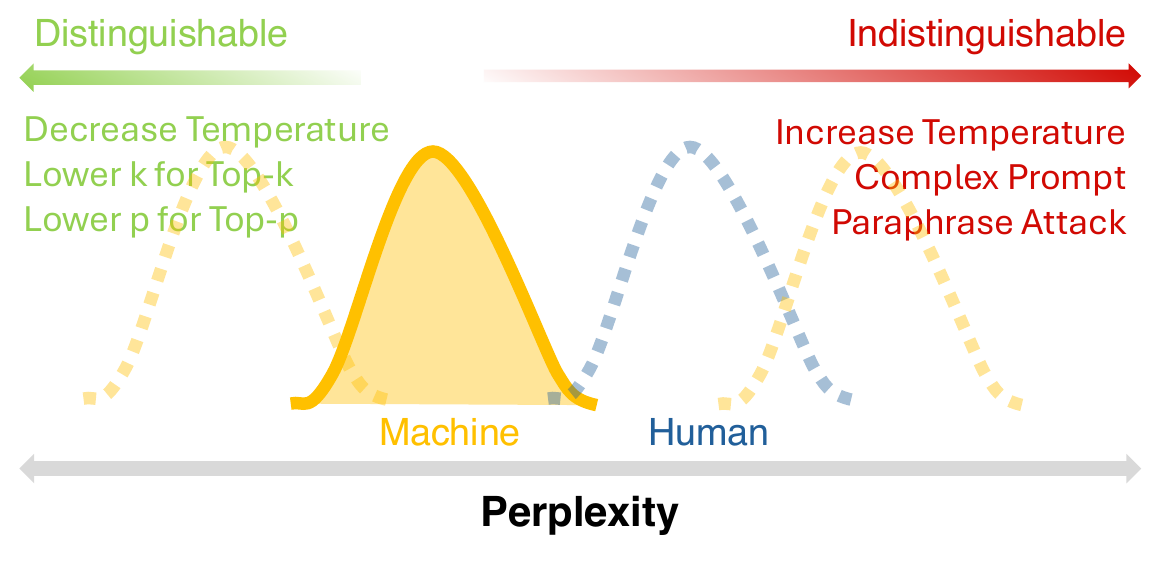}
    \caption{Both logit-based and perturbation-based detectors are not robust to changes in decoding, variations of prompting style, and adversarial attacks.} \vspace{-2mm}
    \label{fig:shift}
\end{figure}

Despite the reported success, the robustness of the above methods are instinctively tied to text distribution, which are influenced by various factors, including but not limited to the following three settings (also summarized in Table \ref{tab:attack_table}): 

    \textbf{1) Decoding Settings}: As is shown in Figure \ref{fig:shift}, \citet{zellers2019defending_fake_news} argues that setting a higher temperature (\textit{e.g.,} T=1.0), a higher k for top-k sampling 
    can increase the likelihood of generating atypical sequences (\textit{i.e.,} increased perplexity), which contradicts the assumption of logit- and perturbation- based methods \cite{ppl_first, vasilatos2023howkgpt}.
    
    \textbf{2) Variations in Prompting}: Changes in prompts can significantly influence the generation process and cause text distribution to be shifted as assessed by the proxy language model, particularly when the prompt is usually unknown to detectors.
    An illustrative example from \citet{binoculars} using the prompt ``Can you write a few sentences about a capybara that is an astrophysicist?'' demonstrates how even seemingly simple prompts can lead to increased perplexity, as the probability that ``capybara being astrophysicist'' is very low. 
   
    In addition, we investigate a planning-based prompting strategy, which emulates human drafting processes and has been widely adopted in neural long-text generation \cite{yao2019plan,tian2022zero,yang2022re3} for improved coherence. Our experiments in Table \ref{tab:main_result} reveal that multiple steps of planning, expansion, and revision (referred to as \uline{complex chains} of prompt, shown in Figure \ref{fig:shift}) results in a significantly higher downstream text perplexity than direct generation (\uline{no chain} of prompt) and one step of planning (\uline{simple chains} of prompt). 
    
    \textbf{3) Paraphrase/Edit Attacks}: Rephrasing a portion of words (termed edit attack) or sentences (termed paraphrase attack) of the original article can dramatically alter the text distribution, too \cite{ghosal2023survey,sadasivan2023can_paraphrase,shi2024redteam}. 
    Such attacks disrupts the original auto-regressive properties, increases output perplexity, and changes text distribution to the indistinguishable region in Figure \ref{fig:shift}.


\section{Machine-Content Detection with Latent Variables}\label{method}

In $\S$ \ref{subsec:hidden_gen}, we formulate the next-token-prediction process with latent variables and introduce a neural model to learn the distributions of these variable. Next, in $\S$ \ref{subsec:combining}, we propose a simple but effective method to combine the benefits of existing sample-space detectors with our latent model.

\begin{table*}[!t]
\centering
\small
\begin{tabularx}{\textwidth}{*3X}
\toprule
\textbf{Sample Space Curvature \qquad +} &
  \textbf{Latent Space PPL \qquad\qquad\qquad $=$} &
  \textbf{Dual Criterion} 
  \\ \midrule

   \hspace{-3mm}\includegraphics[width=1.1\linewidth]{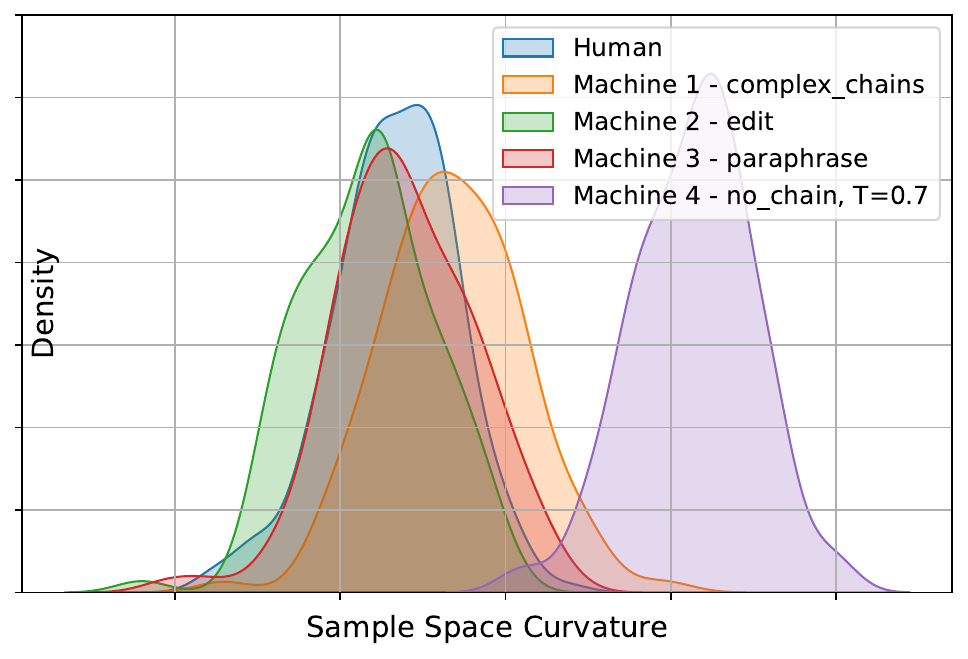}  &
   \hspace{-3mm}\includegraphics[width=1.1\linewidth]{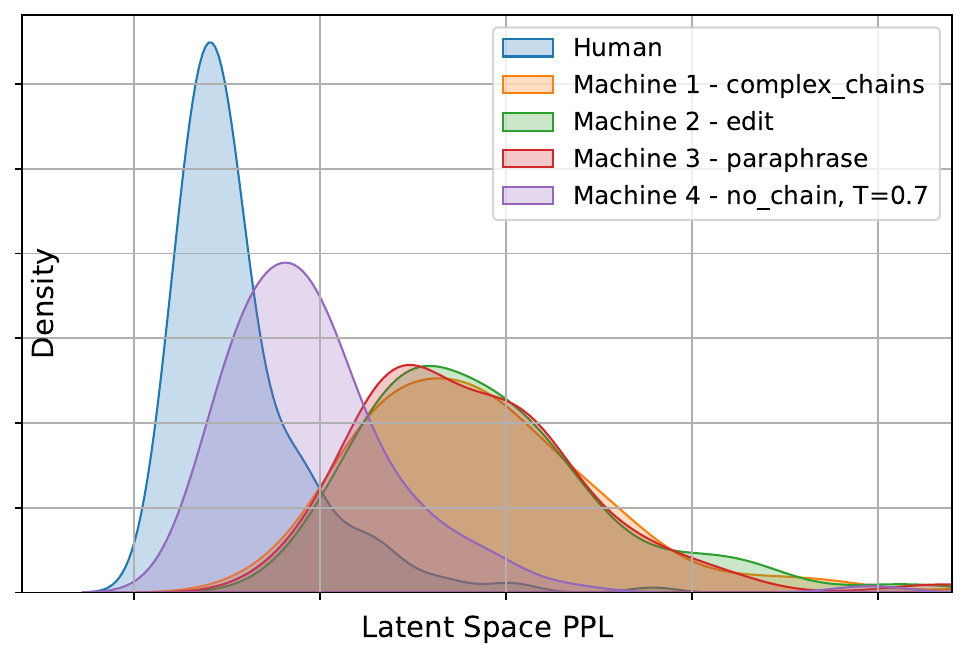} &
   \hspace{-3mm}\includegraphics[width=1.1\linewidth]{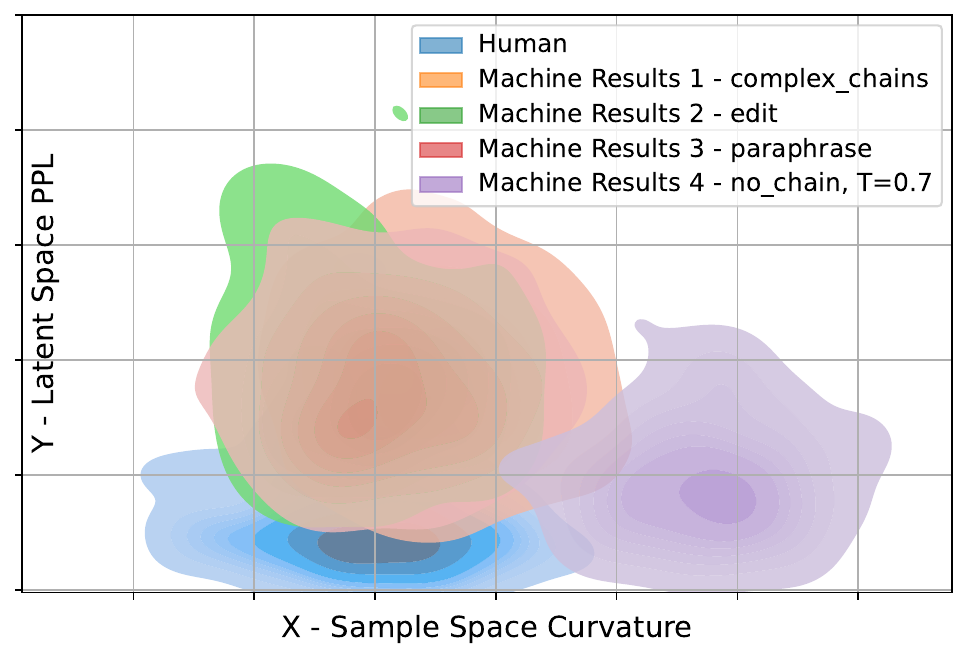}
   \vspace{-5mm}
 \\ \bottomrule
\end{tabularx}
\captionof{figure}{\textbf{Left and middle}: kernel density plots of the sample-space curvature and latent space PPL across five test sets in the news domain. These include \textcolor[HTML]{0D96F3}{human-written} texts (collected from multiple sources) and machine-generated texts under four different configurations. The plots reveal complementary strengths: 1) the sample-space curvature only effectively distinguishes machine outputs generated from \textcolor[HTML]{410DF3}{typical settings} but fail to identify outputs generated with \textcolor[HTML]{F38A0D}{complex prompts} or after \textcolor[HTML]{F67561}{paraphrasing}/\textcolor[HTML]{79D365}{edit} attacks; 2) the latent-space PPL excels at distinguishing those non-standard settings. \textbf{Right}: considering both criteria leads to the most robust detection performance. 
}\label{fig:1D_2D_news} \vspace{-2mm}
\end{table*}

\subsection{Generative Process with Hidden Variables} \label{subsec:hidden_gen}
\paragraph{Formulation}
\begin{figure}[!t]
    \centering
    \includegraphics[width=1\linewidth]{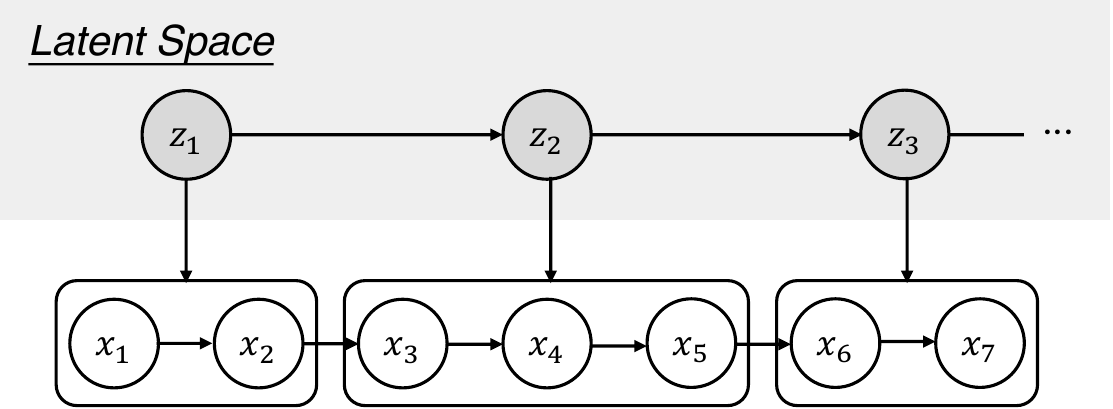}
    \caption{Generative process with latent variables. 
    }\vspace{-2mm}
    \label{fig:latent_model}
\end{figure}

Following \citet{deng2022model_criticism}, we introduce a next-word prediction model that also models underlying structures using latent variables ($z \in \mathcal{Z}$), with the generated output as observed variables ($x \in \mathcal{X}$) shown in Figure \ref{fig:latent_model}:

\begin{equation}
\vspace{-2mm}
\begin{aligned}
    & P(x) = \sum_{z}P(z)P(x \mid z)
\end{aligned}
\end{equation}
\noindent Given a single text sequence $\mathbf{x}$ sampled from $x \sim P_x(x)$, we infer the latent sequence $\mathbf{z}$ from $\mathbf{x}$ with a learned posterior function $P_z(z \mid x)$\footnote{If $P_z(z \mid x)=\mathbbm{1} [z=x]$ or $z$ is the same as $x$, then \textit{Latent PPL} is the same as text perplexity.}, also called a critic model. We can evaluate the negative log-likelihood (latent NLL) and perplexity (Latent PPL) of the inferred latent variables with:

\begin{equation}
\vspace{-2mm}
\begin{aligned}
    \textit{Latent NLL}(\mathbf{x}) & = - \mathbb{E}_{\mathbf{z} \sim P_z(z\mid x)} \log P_z(\mathbf{z})\\
    \textit{Latent PPL}(\mathbf{x}) & = \exp \frac{1}{m} {\textit{Latent NLL}(\mathbf{x})}
\end{aligned}
\end{equation}


\noindent where $m$ is the length of the latent sequence, $\mathbf{z}$. When using an external inference model, the process of inferring latent features is expressed as $z^{\star}=\text{argmax} _z p(z \mid x)$, and the $\textit{Latent NLL}(\mathbf{x})$ becomes $-\log p\left(\text{argmax} _z p(z \mid x)\right)$.

\paragraph{Choice of latent variables} 

We study the impact of the choice of the critic model, $P_z(z \mid x)$. Without losing generality, the latent variables can be anything that captures the high-level underlying structures of a long-form text, such as topic or event transitions. We experiment with five different variables that are relatively easy to obtain, including the part-of-speech tags, nouns or verbs as the approximation of topics, event types, and event triggers, all of which can be obtained using off-the-shelf extraction models \cite{bird2009natural, omnievent}. We show the results of \textbf{\textit{event triggers}} \footnote{\textit{Event type} and \textit{event trigger} are terms commonly used in Information Extraction \cite{grishman1996message, doddington2004automatic, chen2015event, liu2020event}. The \textbf{event trigger} is the main word or phrase that most clearly signals the occurrence of an event, while the \textbf{event type} refers to the category or classification of the event trigger. For example, in the sentence \textit{`Barry Diller on Wednesday quit as chief of Vivendi Universal Entertainment'}, the event trigger is \textit{`quit'}, and the event type is \textit{`Personnel\_End-Position'}.}, the best performing latent variable, in $\S$ \ref{subsec:main_result} and report the full results in $\S$ \ref{subsec:ablation_five_sequences}.

\paragraph{Latent-Space Language Model}
Given a critic model $P_z(z \mid x)$ and a distribution of $x$, we have to learn $P_z(\mathbf{z})$ to obtain \textit{Latent PPL}. Concretely, we train a reduced-scale transformer on sequences of latent variables inferred only from human-written texts. At the test time,  we can then compare the latent PPL from two distributions: human-written and machine-generated.


\begin{algorithm}[t]
\small
\caption{Dual Criterion Process (Sequential)}
\label{algo:sequentially_decide}
\begin{tabular}{ll}
\textbf{Input:} & \\
& $X$ - List of sample space curv. \\
& $Y$ - List of latent space PPL \\
\textbf{Output:} & \\
& combined - List \Comment{requires continuous\\
& criteria rather than binary outcomes.} \\
\textbf{Begin} & \\
0: & \textbf{init} combine as \textbf{empty list}\\ 
1: & \textbf{for} $(x, y)$ \textbf{in} $X, Y$ \textbf{do} \\
2: & \quad \textbf{if} $\text{Confidence} (x=\text{machine})> 95\%$ \textbf{then}\\
3: & \quad \quad combined.add(max($Y$)) \\
4: & \quad \textbf{else} \\
5: & \quad \quad combined.add($y$) \\
6: & \textbf{return} combined\\
\textbf{End} &
\end{tabular}
\vspace{-1mm}
\end{algorithm}

\vspace{-1mm}\subsection{Dual Criterion Process}\label{subsec:combining}
Latent variables contribute to machine-content detection by complementing the strengths of observation-space detectors. We illustrate this by examining the distributions of HWTs and and MGTs under four different configurations in Figure \ref{fig:1D_2D_news}. 
The figure reveals that 1) the sample-space curvature \cite{mitchell2023detectgpt} only distinguishes machine outputs from standard settings (no chain, T=0.7) but fails for those outputs from complex prompts or altered by paraphrasing and editing attacks. Conversely, 2) the latent-space PPL excels at identifying texts generated under these challenging conditions but is less successful in distinguishing machine outputs under typical settings.

Towards more robust detection, we propose to combine both metrics. Follow existing setups \cite{mitchell2023detectgpt,bao2023fastdetectgpt} which report detection accuracy using AUROC\footnote{Area under the True Positive Rate (y-axis) against False Positive Rate (x-axis)}, we require a continuous criterion rather than a mere binary classification outcome. Therefore, we consider both metrics in a sequential order, first based on the confidence level of the sample-space detector. The detailed procedure is described in Algorithm \ref{algo:sequentially_decide}.

\section{Experimental Results}\label{sec:experimental_result}

\subsection{Dataset}\label{subsec:exp_settings}
\paragraph{Human Text}
Following previous setups \cite{bao2023fastdetectgpt,mitchell2023detectgpt},
we collect texts from similar domains that cover a variety of LLM use-cases. We use recent movie synopses from Wikipedia to represent creative writing, New York Times and BBC articles to represent news, and the introduction sections of Arxiv papers from three disciplines: economy, quantitative biology, and computer science, to represent academic essays. We crawled the latest human-written texts ourselves and intentionally avoided using common datasets such as Reddit WritingPrompt \cite{writingprompt}, XSum \cite{xsum} and PubMed \cite{jin2019pubmedqa} to avoid data contamination in recent LLMs such as Llama3 and GPT-4.

\paragraph{Machine Text} On all above domains, we collect machine outputs from two sources: Llama3 \cite{llama3modelcard} that represents open-source LLMs and GPT-4 \cite{gpt4} that represents proprietary models. We test the following and variations in prompts. \uline{\NoChain{}}: Directly generate the output given the task instruction and \textit{generating seed} (including the title, first sentence, topic, etc.). We compare sampling with decoding temperature T=0.7 and T=1.0.
    \uline{\SimpleChain{}}: First write an outline given the task instruction and \textit{generating seed}, then expand the outline to an complete article.
    \uline{\ComplexChain{}}: Identify illogical and vague descriptions and revise the original generation of simple chain. We use a decoding temperature of 1.0 unless otherwise specified.

We also implement popular \textbf{attack} methods on texts generated by complex chain to increase the difficulty of detection task: \uline{\EDIT{}} \cite{shi2024redteam}: Randomly replace 40\% adjectives, adverbials and 20\% verbs (about 15\% in total) with their synonyms. \uline{\PA{}} \cite{sadasivan2023can_paraphrase}: Randomly paraphrase 40\% sentences while maintaining the overall coherence by keeping the proper nouns and writing style unchanged.


\paragraph{Inferring Events as Latent Variable} We extract event types and triggers through OmniEvent's model \cite{omnievent} under the MAVEN schema \cite{wang2020maven} for news and movie. We used GPT-4 for few-shot event extraction in academic essays due to the absence of a specialized model. \footnote{Further details on our data size, source, prompting, attack strategies, and event extraction models are provided in Appendix \ref{appexdix:data}.}

\begin{table*}[!t]
\setlength{\tabcolsep}{4pt}
\centering
\small
\begin{tabular}{l|cc|c|c?cc|c|c?cc|c|c}
\toprule
\multicolumn{1}{c|}{\multirow{3}{*}{ \begin{tabular}[c]{@{}l@{}}Source of\\ Machine Output\end{tabular} }}                        & \multicolumn{4}{c?}{\cellcolor[HTML]{EFEFEF} \textsc{Movie}} & \multicolumn{4}{c?}{\cellcolor[HTML]{EFEFEF} \textsc{News}} & \multicolumn{4}{c}{\cellcolor[HTML]{EFEFEF} \textsc{Arxiv}}\\
\multicolumn{1}{c|}{}                                          & \multicolumn{2}{c|}{\textbf{Sample }} & \textbf{Latent}      & \textbf{Dual} & \multicolumn{2}{c|}{\textbf{Sample }} & \textbf{Latent}      & \textbf{Dual} & \multicolumn{2}{c|}{\textbf{Sample }} & \textbf{Latent}      & \textbf{Dual}   \\ 
 & \textbf{PPL}     & \textbf{Curv.}     & \textbf{PPL} & \textbf{Crit.}& \textbf{PPL}     & \textbf{Curv.}     & \textbf{PPL} & \textbf{Crit.}& \textbf{PPL}     & \textbf{Curv.}     & \textbf{PPL} & \textbf{Crit.}  \\ \midrule
 
1-\ComplexChain{}             &$\xrightarrow{0.84}$ & $\overset{\mathrm{0.53}}{=\joinrel=}$  & $\xrightarrow{0.99}$ & 0.98 &
$\xrightarrow{0.88}$ & $\xrightarrow{0.73}$  & $\xrightarrow{0.97}$ & 0.96&
$\xleftarrow{0.81}$ & $\xrightarrow{0.80}$  & $\xrightarrow{0.78}$ & 0.79 \\

2-\PA{} &$\xrightarrow{0.95}$ & $\xleftarrow{0.65}$  & $\xrightarrow{0.99}$ & 0.99 &
$\xrightarrow{0.97}$ & $\overset{\mathrm{0.53}}{=\joinrel=}$  & $\xrightarrow{0.97}$ & 0.97&
$\xrightarrow{0.53}$ & $\overset{\mathrm{0.52}}{=\joinrel=}$  & $\xrightarrow{0.88}$ & 0.87 \\

3-\EDIT{}  &$\xrightarrow{0.97}$ & $\xleftarrow{0.73}$  & $\xrightarrow{0.99}$ & 0.99 &
$\xrightarrow{0.99}$ & $\xleftarrow{0.62}$  & $\xrightarrow{0.97}$ & 0.97&
$\xrightarrow{0.80}$ & $\xleftarrow{0.73}$  & $\xrightarrow{0.90}$ & 0.88 \\

4-\SimpleChain{}            &$\xrightarrow{0.66}$ & $\overset{\mathrm{0.56}}{=\joinrel=}$  & $\xrightarrow{0.98}$ & 0.97 &
$\xrightarrow{0.70}$ & $\xrightarrow{0.86}$  & $\xrightarrow{0.95}$ & 0.96&
$\xleftarrow{0.91}$ & $\xleftarrow{0.90}$  & $\xrightarrow{0.73}$ & 0.80 \\

5-\NoChain{}                 &$\overset{\mathrm{0.54}}{=\joinrel=}$ & $\xrightarrow{0.75}$  & $\xrightarrow{0.96}$ & 0.95 &
$\xleftarrow{0.88}$ & $\xrightarrow{0.99}$  & $\xrightarrow{0.83}$ & 0.99&
$\xleftarrow{0.90}$ & $\xrightarrow{0.95}$  & $\xrightarrow{0.71}$ & 0.86 \\

6-\NoChain{} \tiny{(T=0.7)}          &$\xleftarrow{0.98}$ & $\xrightarrow{0.99}$  & $\xrightarrow{0.94}$ & 0.99 &
$\xleftarrow{0.94}$ & $\xrightarrow{0.99}$  & $\xrightarrow{0.80}$ & 0.99 & $\xleftarrow{0.98}$ & $\xrightarrow{0.99}$  & $\xrightarrow{0.61}$ & 0.78 \\ \midrule

\rowcolor[HTML]{EFEFEF} 
Mixture                      & 0.642  & 0.646   & \textbf{0.976}   & \uline{0.972}  & 0.565  & 0.750   & \uline{0.912}   & \textbf{0.969} & 0.653  & 0.734   & \uline{0.759}   & \textbf{0.855} \\ \bottomrule
\end{tabular}
\caption[]{\textbf{First 6 rows}: Relative position (indicated by arrows) and the detection accuracy (measured in AUROC) of individual distributions of MGTs and HWTs. \uline{Note that the directions of arrows are more important than the numerical values.} An arrow pointing to the right ($\rightarrow$), left ($\leftarrow$), or equality ($=$) signifies that the machine distribution is to the right of, to the left of, or close to the human distribution. Neither of the sample-space detectors are robust, as is indicated by a mixture of arrows types.  
\textbf{Last row}: Detection accuracy of a mixture of the all distributions of MGTs described above and HWTs, which better reflects real-world black-box scenarios. We use boldface to denote the best performance and underscore the second best.
} 
\label{tab:main_result}
\end{table*}


\begin{table*}[!t]
\centering
\small
\begin{tabularx}{\textwidth}{*3X}
\toprule
\textbf{Movie Synopsis} &
  \textbf{News} &
  \textbf{Scientific Writing} 
  \\ \midrule

   \hspace{-3mm}\includegraphics[width=1.1\linewidth]{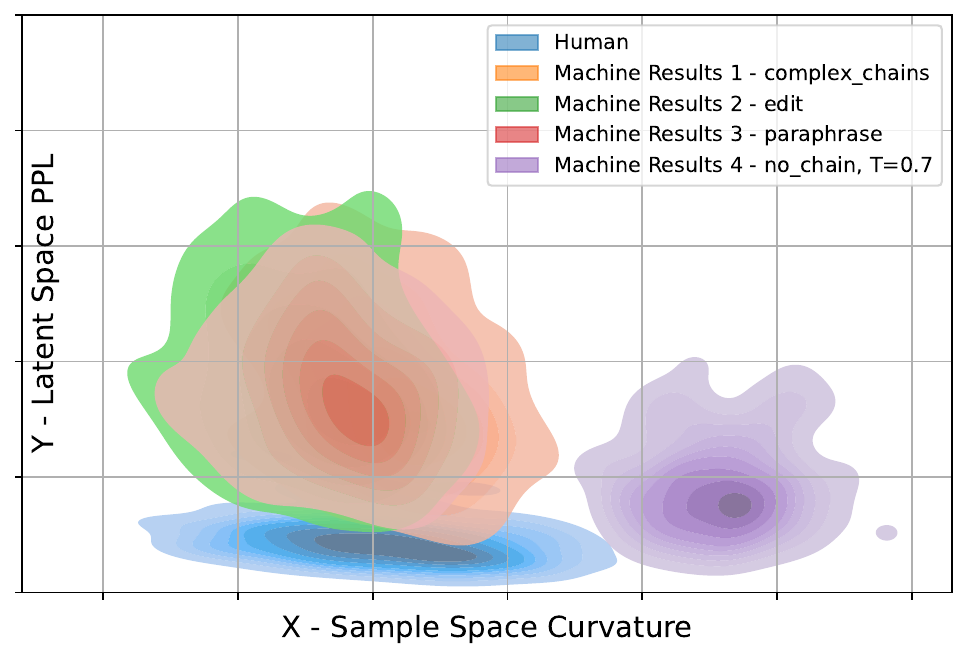}  &
   \hspace{-3mm}\includegraphics[width=1.1\linewidth]{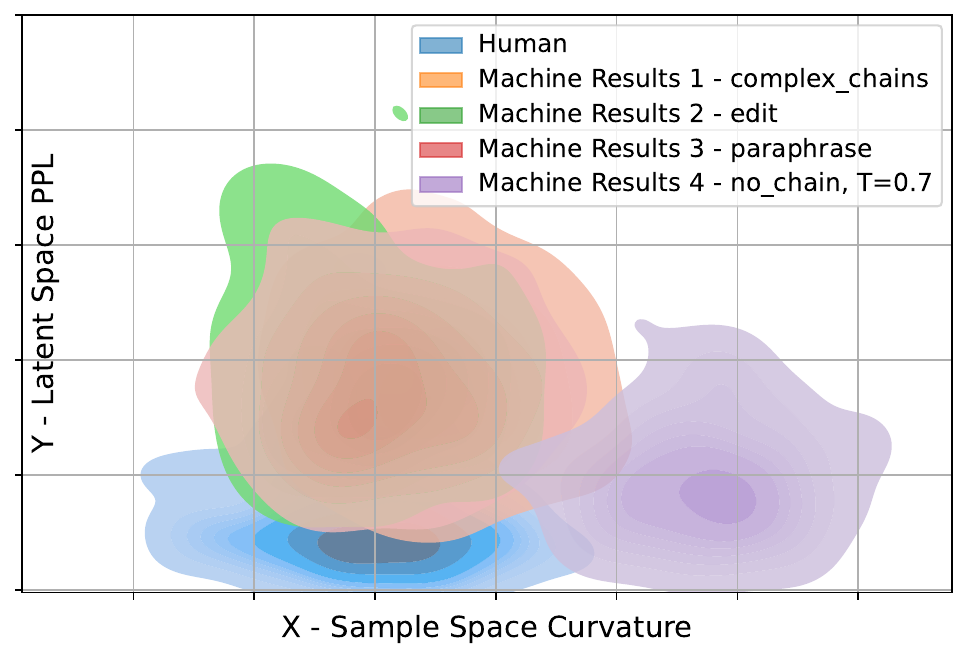} &
   \hspace{-3mm}\includegraphics[width=1.1\linewidth]{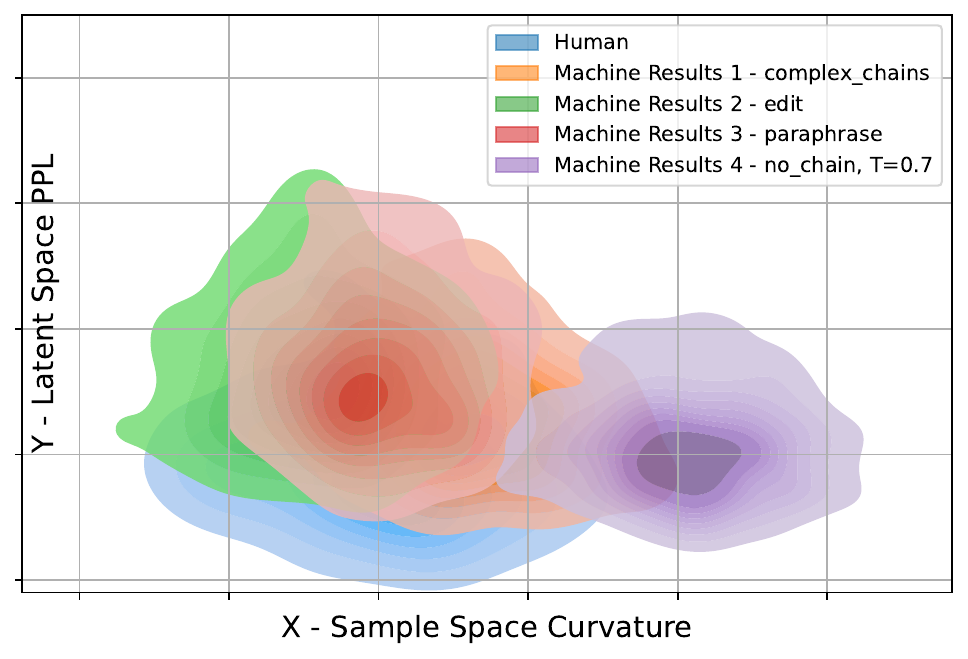}
   \vspace{-3mm}
 \\ \bottomrule
\end{tabularx}
\captionof{figure}{2D density clouds. For better readability, we only show four sets of machine generated outputs.
}\label{tab:2D_cloud_all} \vspace{-2mm}
\end{table*}

\subsection{Baseline Methods}

We compare our method with the state-of-the-art detection system, Fast-DetectGPT \cite{bao2023fastdetectgpt}, which utilizes the conditional probability curvature (\textbf{Sample Curv.}) using a GPT-Neo-2.7b model \cite{gpt-neo}. It outperforms the well-known DetectGPT \cite{mitchell2023detectgpt} by 29\%. In addition, we compare with the popular token perplexity (\textbf{Sample PPL}) on several pretrained LLMs including gpt2-medium, gpt2-large \cite{radford2019language}, GPT-Neo-2.7b, and Llama3-8b \cite{llama3modelcard}, and find similar accuracy despite of model sizes. To be consistent with the model used in \textit{Sample Curv.}, we picked GPT-Neo-2.7b.

\subsection{Training Configurations}
\paragraph{Latent Model} Since our latent sequences are much shorter than the observed texts, we train a lightweight transformer \textit{from scratch} on sequences of latent variables inferred solely from HWTs. We began with a gpt2-medium-sized transformer and performed grid search to decrease parameter size by half every iteration. We continued to reduce the model size until a noticeable decrease on a held-out development set was observed. The model configuration is a randomly initialized transformer with 12 heads, 12 layers, and 384 embedding dimensions.
We still consider our detector as ``zero-shot'' as it is not trained on negative samples at all.

There was not much effort needed to tune the hyper-parameters. We trained our latent-space model with a learning rate of 1e-4, batch size of 24, and drop-out rate of 0.1. We randomly split 10\% from the training set as a validation set. We stopped training when the model reached the maximum number of epochs, which is set to 10. We selected the model checkpoint with the best validation loss and ran that model on our test set.

\paragraph{Domain Adaptation}
To ensure a fair comparison, we performed domain adaptation (DA) by fine-tuning all sample-space detectors for one epoch on the same dataset of human texts on which our latent model is trained. However, we found that such domain-adaptation resulted in decreased performance at test time, possibly due to over-fitting. Consequently, we used the pre-trained LMs without DA and report their detection accuracy.



\subsection{Main Results}\label{subsec:main_result}

We report the detection performances of all compared models in Table \ref{tab:main_result}. We first evaluate their accuracy on six individual sets of MGTs and HWTs (the first six rows in Table \ref{tab:main_result}), and visualize their distributions in Figure \ref{tab:2D_cloud_all}. Across the three domains of movie, news, and science, neither of the sample-space detectors demonstrate robustness to all generation configurations or attacks, as is reflected by the variability in arrow directions—right, left, and equal. Consequently, when facing a mixture of all six sets of MGTs that better reflects the real-world black-box settings (the last row in Table \ref{tab:main_result}), sample-spaces detectors achieve only 60\% to 70\% accuracy. On the other hand, our latent-space detector consistently place machine outputs in a separable space (\textit{i.e.,} consistently to the right of human distributions), greatly surpassing the baselines. Our dual criterion process which takes advantage of both the sample-space curvature and \textit{latent PPL} achieves the highest detection accuracy. 

\begin{table*}[!t]
\small
\centering
\begin{tabular}{@{}cc|ll|ll|ll@{}}
\toprule
\multicolumn{2}{c|}{\textbf{Length of}}  & \multicolumn{2}{c|}{\textbf{Movie}} & \multicolumn{2}{c|}{\textbf{News}} & \multicolumn{2}{c}{\textbf{Arxiv}} \\ 
\textbf{Token} & \textbf{Event Trigger} & \textbf{S Curv.} & \textbf{L PPL} & \textbf{S Curv.} & \textbf{L PPL} & \textbf{S Curv.} & \textbf{L PPL} \\ \midrule
128 & 16.4 & 0.567 & \textbf{0.829} & 0.562 & \textbf{0.842} & \textbf{0.605} & 0.549 \\
256 & 28.7 & 0.579 & \textbf{0.916} & 0.604 & \textbf{0.887} & \textbf{0.652} & 0.586 \\
512 & 55.9 & 0.618 & \textbf{0.966} & 0.667 & \textbf{0.909} & \textbf{0.729} & 0.679 \\
1024 & 83.3 & 0.636 & \textbf{0.976} & 0.738 & \textbf{0.912} & 0.734 & \textbf{0.767} \\
2048 (max) & 84.2 & 0.646 & \textbf{0.976} & 0.750 & \textbf{0.912} & 0.734 & \textbf{0.768} \\ \bottomrule
\end{tabular}
\caption{Detection accuracy (AUROC) of the strongest sample-space detector (\textbf{S Curv.}, \cite{bao2023fastdetectgpt}) and our latent-space detector (\textbf{L PPL}) with varying content lengths. Higher accuracy for each domain is highlighted in bold.
}
\label{table:length}
\end{table*}

We also observe that the latent-space model demonstrates superior performance in narrative domains (\textit{e.g.,} movie and news) compared to scientific domains. This can be attributed to two factors. \textbf{First}, narratives fundamentally rely on events as their central structural elements \cite{verhoeven2004relating,keven2016events}, whereas scientific writing focuses more on factual information and technical descriptions, which may not align as perfectly with the event-centric nature of our latent representation. \textbf{Second}, we find the current event extraction model less reliable on scientific texts. This limitation can lead to error propagation during both training and testing phases. 
Therefore, we encourage future research to develop specialized methods for extracting alternative discourse structures from academic writings, which may improve the accuracy of detection in scientific domains.

\subsection{Impact of Input Sequence Length}

The detection accuracy of both our latent-space detector and the token-level baselines is impacted by the length of text. To account for this, we conduct additional experiments controlling for input length by truncating texts to \{128, 256, 512, 1024, 2048\} tokens and extract event triggers. We compare the performance of the strongest sample-space detector with our latent-space detector and report their test accuracy in Table \ref{table:length}. For both methods, detection accuracy improves with longer text. Notably, despite our approach relying on discourse features, which are sparser than tokens, it demonstrates greater robustness to shorter texts, particularly in the movie and news domains.

\section{Further Analysis}\label{sec:analysis}
 \begin{table}[]
 \centering
 \setlength{\tabcolsep}{4pt}
 \small
\begin{tabular}{@{}lccccc@{}}
\toprule
                     & \textsc{Movie} & \textsc{News}  & \textsc{Arxiv} & \textbf{Avg} & \textbf{Relative}\\ \midrule
Sample Curv. & 0.65 & 0.77 & 0.73 & 0.72 & - \\ \midrule
Pos Tag & 0.53 & \uline{0.79} & 0.65 & 0.66 & - 8\% \\
Verbs & 0.80 & 0.64 & \textbf{0.78} & 0.74 &+ 4\%\\
Nouns & 0.78 & 0.51 & 0.66 & 0.65 &- 9\%\\
Event Type & \uline{0.85} & 0.75 & 0.73 & \uline{0.78} &+ 9\%\\
\rowcolor[HTML]{EFEFEF} Event Trigger & \textbf{0.98} & \textbf{0.91} & \uline{0.77} & \textbf{0.89} &+ 24\%\\ \bottomrule
\end{tabular}
\caption{Detection accuracy (measured in AUROC) and relative performance change using models trained on different latent variables. We highlight the best in boldface and second best in underline. For \textsc{Arxiv}, we posit that the higher accuracy with verb sequences is due to the errors of events extracted from scientific writings.} \vspace{-2mm}
\label{tab:best_choice_variavles}
\end{table}

\begin{table*}[!t]
\centering
\small
\begin{tabularx}{\textwidth}{*3X}
\toprule
\textbf{Movie} &
  \textbf{News} &
  \textbf{Arxiv} 
  \\ \midrule

   \includegraphics[width=1.\linewidth]{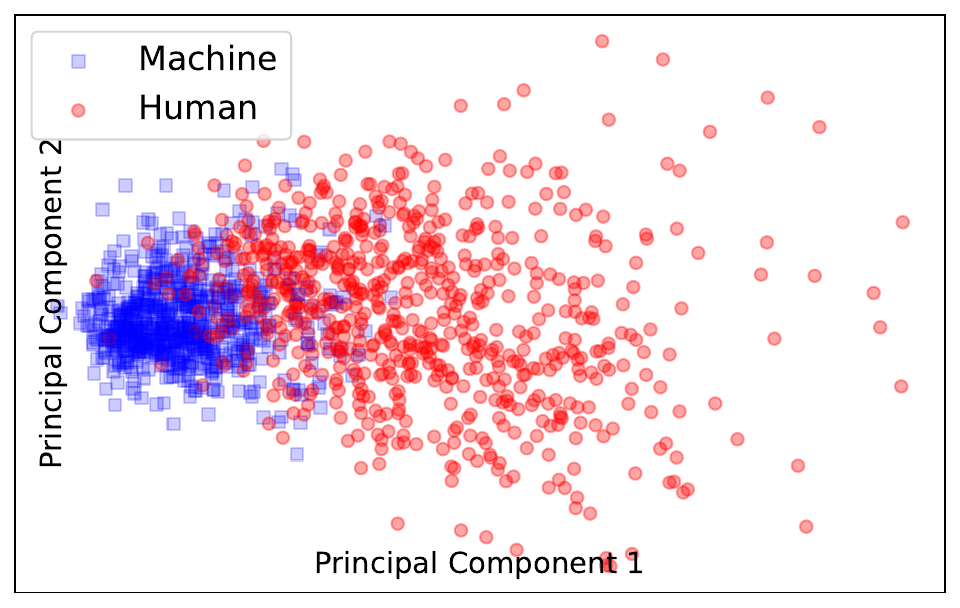}   &
   \includegraphics[width=1.\linewidth]{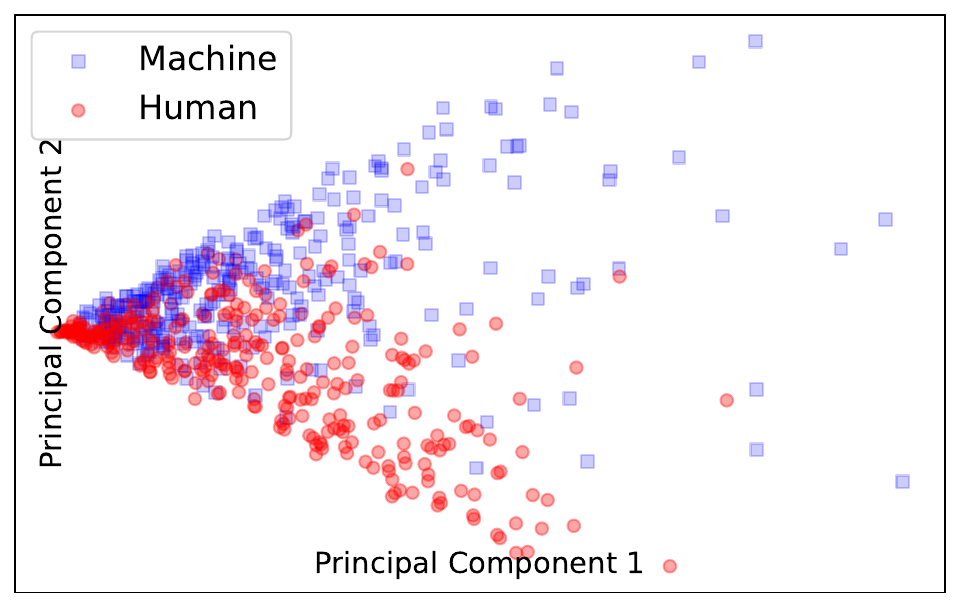} &
   \includegraphics[width=1.\linewidth]{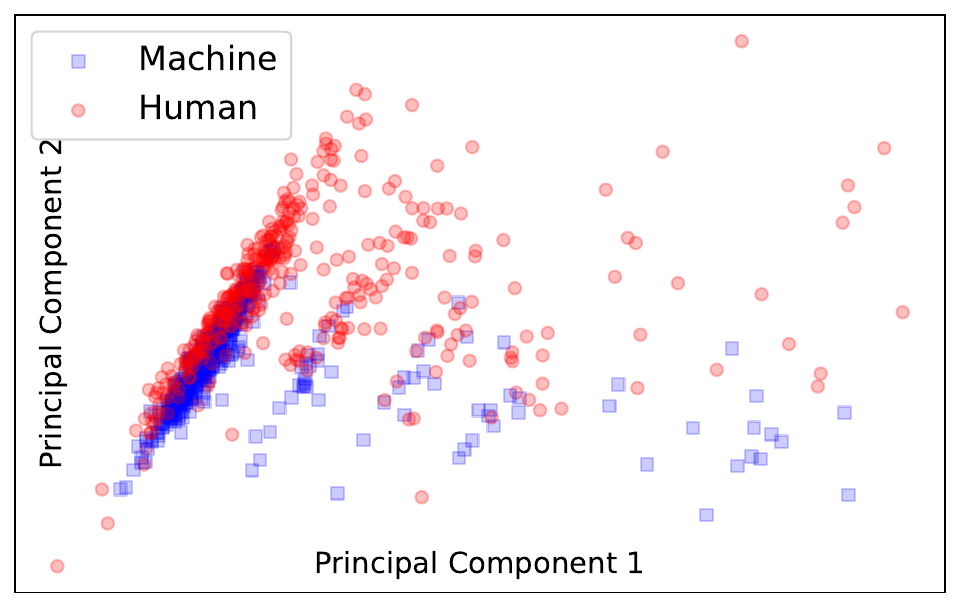} 
 \\ 
 \bottomrule \vspace{-2mm}
\end{tabularx} \vspace{-5mm}
\captionof{figure}{Bag-Of-Words feature of human and machine latent variables, reduced via PCA.}\label{figure:PCA_latent_variables}\vspace{-2mm}
\end{table*}

\begin{table}[]
\centering
\small
\begin{tabular}{@{}llll@{}}
\toprule
 & \multicolumn{1}{l}{\textbf{Movie}} & \multicolumn{1}{l}{\textbf{News}} & \multicolumn{1}{l}{\textbf{Arxiv}} \\ \midrule
\textbf{BOW - PC}       & 0.945 & 0.731 & 0.705 \\
\textbf{Sequence} $_{\text{Sf.}}$ & 0.953 & 0.801 & 0.744 \\
\rowcolor[HTML]{EFEFEF} \textbf{Sequence}       & 0.976 \tiny{($\uparrow$3.3\%)}  & 0.912 \tiny{($\uparrow$24.8\%)} & 0.768 \tiny{($\uparrow$8.9\%)}\\ \bottomrule
\end{tabular}
\caption{Detection accuracy (measured in AUROC) of two latent methods that only consider the choices of event triggers (BOW-PC and Sequence $_{\text{Sf.}}$) and our best latent model that includes both the choices and transitions (Sequence).}
\vspace{-3mm}
\label{tab:sequence_ablation}
\end{table}

\subsection{What is the best choice of latent variables?}\label{subsec:ablation_five_sequences}

We report the detection accuracy of five different latent variables: part-of-speech tags, nouns or verbs, event types, and event triggers in Table \ref{tab:best_choice_variavles}. We find that using parts of speech (which represents the inner-sentence coherence) as the underlying hidden structures leads to decreased accuracy. This also supports the claim that \uline{current LLMs already generate locally human-like texts}. On discourse structures, \uline{only events are indicative of the deviations} observed between LLMs and humans. Event types, ranking as the second most effective indicator, provide a richer analysis beyond mere word forms (\textit{e.g.} verbs and nouns) and syntactic functions (\textit{e.g.} parts of speech). However, event types might be too generic and cannot capture the finer semantic differences as well as event triggers.


\subsection{Event choice v.s. Event transition}

A natural follow-up question is \textit{``How much difference comes from event choices versus event transitions?''}
To answer this, we build Bag-Of-Words (BOW) models using the event triggers extracted from the test set, and visualize the PCA-reduced features in Figure \ref{figure:PCA_latent_variables}. We compare the detection performance using three methods in Table \ref{tab:sequence_ablation}: 1) the principle component of the BOW features (BOW-PC), 2) randomly shuffled event triggers (Sequence $_{\text{Sf.}}$), and 3) ordered event triggers (Sequence).

\paragraph{Event transition is crucial, but the extent depends on task creativity.} When event triggers are randomly shuffled or reduced to their principle component, the loss of sequential information impairs the detector's ability to distinguish text origins. This highlights the critical role of \uline{event transitions} in maintaining an article's coherence and authenticity. However, the amount of decrease are still domain dependent. In movies, the choice of events itself already plays a decisive role (over 94\%). This suggest that LLMs use a distinct list of events triggers from humans in highly creative writing tasks such as movie.
Conversely, in less open-ended domains like news and science, the choice of event triggers is less distinct and the transition between event triggers contributes more to authenticity. This indicates that while LLMs have learned the appropriate event triggers, they have not yet mastered the most logical flow of event transitions. To sum up, MGTs are farther away from HWTs on highly creative generation tasks.




\subsection{Does explicit structure-aware planning make our detector less effective?}

We utilize the differences of latent-structures between MGTs and HWTs. If models are explicitly instructed to elaborate on the underlying structures before auto-regressive generation, would our detection method remain effective? The two prompting methods (\textit{i.e.,} simple chain and complex chain) we employed in $\S$ \ref{subsec:exp_settings} are designed to answer this question by integrating a preliminary planning and revision stage before or after text generation.

We emphasize that \uline{current models are unable to mimic human-like discourse authenticity even when instructed to plan on these structures.}
A comparison of line 1, 4, and 5 in Table \ref{tab:main_result} reveals that despite the addition of a planning and revision stage, models still struggle to replicate the human-like flow in event arrangement. This finding echos previous critiques that current LLMs are poor in mimicking human high-level structures \cite{deng2022model_criticism}. One possible reason is that most LLMs are trained to optimize for local coherence and fluency, rather than an overarching, discourse-level logic. For example, the planning mechanism employed by LLMs usually involve skeletal outlines or lists of keypoints that tend to prioritize surface-level coherence instead of the depth in thought. In contrast, human writers often plan their articles with a conscious awareness of theme and plot development that are inherently challenging for current LLMs. 
\section{Related Work}

\subsection{MGT Detection}
In aspects of ownership and usability, detectors can be roughly divided into \textit{a priori} and \textit{post-hoc} categories. \textbf{A priori method} involves proactive involvement of the model's generation process through techniques like watermarking \cite{ghosal2023survey}. For example, \citet{watermark} encourage the sampling of tokens from a pre-determined category (a green list), and this special token distribution can be utilized for detection. \citet{christ2023undetectable_watermark} minimizes the distance between the watermarked and original distribution, making the watermark both undetectable and unbiased in expectation. Despite the effectiveness of such techniques, users may still opt for non-watermarked model like GPT-4, underscoring the need for robust \textit{post-hoc} detection methods \cite{detect_survey2}. 
\textbf{Post-hoc method} involves fine-tuning classifiers on corpora of positive and negative samples (HWTs and MGT) \cite{gpt_sentinel, liu2019roberta} and zero-shot detection. These former classifiers often struggle with out-of-distribution data \cite{zhang2023assaying} and are sensitive to data quality \cite{liang2023gpt_bias_non}. The variety of existing LLMs makes it less practical to curate a universal training dataset \cite{bhattacharjee2023conda}. The zero-shot detectors perform the task through a statistical approach. As is introduced in $\S$ \ref{sec:fragility}, \citet{su2023detectllm, gehrmann-etal-2019-gltr, mitchell2023detectgpt, bao2023fastdetectgpt} employ statistics like probability, entropy and curvature. We focus on zero-shot detection because it is generalizable to diverse domains and do not require access to the source model.

\subsection{Attacks to Zero-Shot Detection}
Attacks can occur pre-, post-, or during text generation \cite{wang2024stumbling}, which we considered in our experiments ($\S$ \ref{subsec:exp_settings}). Pre-generation attacks involve manipulating prompts to produce outputs that are inherently harder to detect, such as adversarial searches to known detectors \cite{shi2024redteam}. Post-generation attacks 
replace text segments with lexically or semantically similar alternatives, such as typos, filled mask, synonyms, and rephrased sentences \cite{shi2024redteam, sadasivan2023can_paraphrase}. On-generation attacks \cite{wang2024stumbling} involve decoding with intentional perturbations like typos or emojis, which are later removed to alter the text's statistical distribution, impairing detection performance. Additionally, \citet{zhang2023assaying} explores how shifts in topic can impact detector efficacy.

\subsection{Latent Features for Language Modeling}
Current LLMs excel at generating locally fluent sentences, yet they often fail to maintain the long-form coherence, which requires awareness of connecting diverse ideas logically \cite{lin2020limitations_autoregressive}. \citet{bowman2015generating_cont} introduced latent variable models to improve the structural understanding of long texts. Following this, Contrastive Predictive Coding (CPC) \cite{oord2018representation_cpc} was proposed to learn unconditioned latent dynamics implicitly, which \citet{wang2022language_stochastic} further refined with the introduction of Brownian bridge to impose structured, goal-oriented dynamics within the latent space of texts. Novel evaluations for long-form coherence include model criticism based on latent structures such 
as section labels in Wikipedia \cite{deng2022model_criticism}. \citet{sheng2024bbscore} created a coherence assessment metric grounded in Brownian bridge theory \cite{horne2007analyzing}. Owning to the lack of reliable methods for inferring completely unobservable features from texts, we constrain our latent variables to more accessible ones such as events.


\section{Conclusion and Discussion}
 
 We propose a novel zero-shot detection framework that employs latent features such as sequences of events. Our method leverages the limitations of current LLMs in replicating authentic human-like discourse, despite their ability to generate locally convincing language. 
 Experimental results demonstrate that our detector is highly robust across various real-world generation settings and attacks.

Similar to existing works, our detection methods also rely on certain assumptions about how machine-generated content differs from human-created content. We specifically assumes that text generators struggle to replicate human-like high-level structures. However, we believe that it will take longer for LLMs to catch-up with humans on high-level structures than for them to mimic token-level distributions. Therefore, our method is likely to remain relevant for a longer period than the token-level approaches. Furthermore, we hope the latent-space statistics can serve as a valuable indicator to guide current LLMs in improving high-level content generation.

\section*{Acknowledgements}

The authors are grateful to Tao Meng for his valuable feedback on the experiments. We also thank the Pluslab members at UCLA and the anonymous reviewers for their insightful comments on the paper. 
This research is partly supported by National Science Foundation CAREER award \#2339766, an Amazon AGI foundation research award, and a Google Research Scholar grant. 
\section*{Limitations} 
Our approach involves inferring discourse features, which are more sparse than tokens, hence is  restricted to detecting long-form texts. Additionally, the detection accuracy is reliant on the performance of an external inference model, which, in our case, is the event extractor. We find the existing event extraction models are less accurate on scientific texts, which can lead to error propagation during both training and testing phases.
We also encourage future research to develop specialized methods for extracting alternative discourse structures from academic writings, which could enhance the accuracy of machine detection in scientific domains.

\newpage
\bibliography{anthology,custom}
\newpage
\appendix

\clearpage
\section{Datasets}\label{appexdix:data}

\begin{table*}[t!]
\small
\centering
\begin{tabular}{lllllll}
\rowcolor[HTML]{EFEFEF} 
\toprule
\textbf{Main Tasks} & \textbf{Source} & \textbf{\begin{tabular}[c]{@{}l@{}}Event Extraction\\ Model\end{tabular}} & \textbf{\begin{tabular}[c]{@{}l@{}}Length\\(Words) \end{tabular}} & \textbf{\begin{tabular}[c]{@{}l@{}}Length\\(Events) \end{tabular}} & \textbf{Train Size} & \textbf{Test Size}\\ \midrule
\begin{tabular}[c]{@{}l@{}}Creative\\ writing\end{tabular} & \begin{tabular}[c]{@{}l@{}}Movie synopsis\\ (from Wikipedia)\end{tabular}  & \begin{tabular}[c]{@{}l@{}}OmniEvent\\\cite{omnievent}\end{tabular} & 673 & 86 & 2,178 & 250\\
\rowcolor[HTML]{EFEFEF} News & NYT and BBC & OmniEvent & 1024 (truncated) & 92 & 2,813 & 218 \\
\begin{tabular}[c]{@{}l@{}}Academic\\ essay\end{tabular}   & \begin{tabular}[c]{@{}l@{}}Arxiv (introduction \\ of computer science,\\economy, and\\quantitative bio)\end{tabular} & Few-Shot GPT-4 & 796 & 66 & 2,680 & 287 \\ \bottomrule
\end{tabular}
\caption{The statistics of our human data. Note that for news articles, word length is truncated to max sequence length of GPT-2.
}
\label{tab:dataset_summary}
\end{table*}

We describe how we created human and machine text dataset in more detail. The statistics summary of dataset is shown in Table \ref{tab:dataset_summary}.

\subsection{Collecting Human Texts}
Considering any existing text from internet can be the training data of current LLMs, we crawled the latest texts to avoid LLMs memorizing them when performing detection. 

For \textbf{Movie}, we crawl the recent English-language films category on Wikipedia\footnote{https://en.m.wikipedia.org/wiki/Category:2020s\_English-language\_films}. To increase the quality of synopses, we remove those with fewer than 25 sentences. To minimize the risk of model memorization, we filter out well-known movies using the lengths of Wikipedia pages as an approximate indicator of popularity.

For \textbf{News} articles, we collected all news articles from the main page of The New York Times\footnote{www.nytimes.com/section/us} published from 2024-04-09 to 2024-05-18 covering mainly politics, business and editorials news. The test data are a mix of New York Times from the same source and BBC\footnote{https://www.bbc.com/news/us-canada}, the latter of which we consider as out-of-distribution to increase the task difficulty.

For \textbf{Arxiv} paper, we downloaded 419 economy, 666 quantitive biology and 782 computer science published from 2023-06 to 2024-04 using its official API. We then extracted the introduction section in plain text from TeX source code of each paper using GPT-3.5.

\subsection{Collecting Machine Texts}
All human and machine outputs have roughly the same length. For each human text in test set, we generate a paired machine-generated text.
\paragraph{Variations in Prompts} For whole text generation, we use pure sampling at temperature T = 1.0 as default.  To further avoid data contamination, we first let the model to rephrase the titles and initial settings, termed as \textit{generating seed}, by altering all the unique identifiers such as proper nouns. 
Then, we use the three prompting strategies described in $\S$ \ref{subsec:exp_settings} to collect machine generated texts from the \textit{generating seeds}. More concretely, \uline{\NoChain{}} directly complete the whole texts.
\uline{\SimpleChain{}} mimics the human-like plan-write process, by first generating an structured outlines and then expanding it to the whole texts. \uline{\ComplexChain{}} add revision steps on top of \uline{\SimpleChain{}}, to add more details in the outline and fix any illogical and vague descriptions in original output. The complex chain prompt used to generate scientific essays is shown in Figure \ref{code:complex_chain}.


\paragraph{Attack} For both paraphrasing and edit attacks, 
we introduce adversarial searches to known detectors \cite{shi2024redteam}. Overall, our approach first generates multiple substitutions for all candidate segments that can be replaced by substitutions without changing the meaning drastically. Then we randomly sample substitutions with certain probability to produce candidates. Finally, GPT-2-XL is used to calculate and select the text with the highest perplexity to gain the maximum attack efficiency at a fixed replacement ratio.  For \textbf{Paraphrase}, the segment is sentence level. We generate 2 to 5 substitutions for each sentence while keeping every proper noun and overall writing styles. Then, we generate candidates through replacing 40\% original sentences. For \textbf{Edit}, the segment is word level. We generate 2 to 5 context-based synonyms for each adjective, adverbial and verb as replacing them would not affect the semantics severely. Then, we generate candidates through replacing 40\% adjectives, adverbials and 20\% verbs (about 15\% of total words). The prompt used for attacks is shown in Figure \ref{code:attack}.

\subsection{Event annotation}
For news and movie, we employed the off-the-shelf T5 model from OmniEvent \cite{omnievent}, which is trained on multiple dataset including ACE05\footnote{https://catalog.ldc.upenn.edu/LDC2006T06}, MAVEN \cite{wang2020maven}, etc. We use this model under MAVEN schema, which defines 168 event types that cover various general scenarios useful for our analysis. Owning to the fact that there are no event extraction model specialized in scientific writing domain, we prompt GPT-4 to extract, as is shown in Figure \ref{code:even_annotation}. Additionally, we employ SpaCy's lemmatization pipeline to standardize the form of all event triggers.


\begin{figure*}[t]
\begin{lstlisting}
<-- Create Consice Outline -->
User: Create a simple outline structure for writing the introduction section of an academic paper based on the given title and first sentence. Each line is a key component and its explanation. The paper domain is "{domain}", title is "{rephrased_title}" and first sentence is "{rephrased_first_sentence}".

Assistant:  "{concise_outline}"

<-- Expand Outline -->
User: Based on the given simple outline of writing the introduction section of an academic paper, expand on the key points outlined, providing bullet points of clear, well-developed arguments, data, context, etc. that strengthen the introduction. The paper domain is "{domain}", title is "{rephrased_title}" and the outline is "{concise_outline}".

Assistant:  "{expanded_outline}"

<-- Draft Paper -->
User: Build upon the given bullet points to write a comprehensive and logically structured introduction that frames the paper's arguments and significance. Your output should be the introduction section of an academic paper generated in about 50 sentences. The paper domain is "{domain}", title is "{rephrased_title}" and the outline is "{expanded_outline}".

Assistant:  "{paper_draft}"

<-- Refine Paper -->
User: Based on the given outline, reexamine the flow of the draft introduction to ensure that it logically progresses from general context to specific research questions, effectively setting up the research framework. Strengthen transitions between ideas, ensure coherence in the presentation of arguments, and align the structure with academic standards for introductions. Your output should be the introduction section of an academic paper generated in about 30 sentences. The paper domain is "{domain}", title is "{rephrased_title}", the given outline is "{expanded_outline}" and the draft is "{paper_draft}". 

Assistant:  "{refined_draft}"

\end{lstlisting}
\caption{Complex chain prompt used for generating scientific essays.}
\label{code:complex_chain}
\end{figure*}

\begin{figure*}[t]
\begin{lstlisting}
<-- Edit Attack -->
User: Given the sentence and words within, for each of words, given two to five substitution words that do not change the meaning of the sentence. Only generate substutions when a word is general but not proper word. Return each general word and its substitutions in one line, in the format of 'word: substitution 1, substitution 2, ...'. sentence: "{sentence}"; words: "{words}"

Assistant:  {Word substituions}

<-- Paraphrase Attack -->
User: Please paraphrase the highlighted sentence (wrapped by '**' ) in the below text in 2 - 5 ways. You should keep all proper words and style of the original text in your paraphrased sentences. Your should directly output paraphrase splitted by linebreak without '**'.\n\nText: "{text}"

Assistant:  {Sentence substitutions}
\end{lstlisting}
\caption{Prompt for edit and paraphrase attack.}
\label{code:attack}
\end{figure*}

\begin{figure*}[t]
\begin{lstlisting}
User: 
Task: For each sentence in input, extract all the major event triggers. Your output should only be a valid JSON string that is a list of dictionary. Each dictionary contains two fileds: 'sentence' and 'triggers'.

Examples: "{examples}"

Now extract all major event triggers in the following input: "{input_sentence}"

Assistant:  {extracted_events}
\end{lstlisting}
\caption{Prompt for event annotation using GPT-4.}
\label{code:even_annotation}
\end{figure*}
\label{sec:appendix}

\end{document}